\documentclass[runningheads]{llncs}

\usepackage{amssymb}
\usepackage{graphicx}
\usepackage{epsfig}

\usepackage{url}
\urldef{\mailsa}\path|{alfred.hofmann,ursula.barth,ingrid.beyer,natalie.brecht,|
\urldef{\mailsb}\path|christine.guenther,frank.holzwarth,piamaria.karbach,|
\urldef{\mailsc}\path|anna.kramer,erika.siebert-cole,lncs}@springer.com|
\newcommand{\keywords}[1]{\par\addvspace\baselineskip
\noindent\keywordname\enspace\ignorespaces#1}

\begin{document}

\mainmatter  

\title{Direct attacks using fake images\\ in iris verification}

\author{Virginia Ruiz-Albacete\and Pedro Tome-Gonzalez \and Fernando Alonso-Fernandez \and Javier Galbally \and
Julian Fierrez \and Javier Ortega-Garcia}
\authorrunning{BIOID 2008}

\institute{Biometric Recognition Group - ATVS \\ Escuela
Politecnica Superior - Universidad Autonoma de Madrid \\
Avda. Francisco Tomas y Valiente, 11 - Campus de Cantoblanco \\
28049 Madrid, Spain - \url{http://atvs.ii.uam.es} \\
\email{\{virginia.ruiz, pedro.tome, fernando.alonso,
javier.galbally, julian.fierrez, javier.ortega\}@uam.es}}

%
%

\toctitle{Lecture Notes in Computer Science}
\tocauthor{Authors' Instructions}
\maketitle

\begin{abstract}
In this contribution, the vulnerabilities of iris-based
recognition systems to direct attacks are studied. A database
of fake iris images has been created from real iris of the
BioSec baseline database. Iris images are printed using a commercial
printer and then, presented at the iris sensor. We use for our experiments
a publicly available iris recognition system. Based on
results achieved on different operational scenarios, we show
that the system is vulnerable to direct attacks, pointing out the importance
of having countermeasures against this type of fraudulent actions.
\end{abstract}

\section{Introduction}

The increasing interest on biometrics is related to the number of
important applications where a correct assessment of identity is a
crucial point. The term \emph{biometrics} refers to automatic
recognition of an individual based on anatomical (e.g., fingerprint,
face, iris, hand geometry, ear, palmprint) or behavioral
characteristics (e.g., signature, gait, keystroke dynamics)
\cite{[Jain06]}. Biometric systems have several advantages over
traditional security methods based on something that you know
(password, PIN) or something that you have (card, key, etc.). In
biometric systems, users do not need to remember passwords or PINs
(which can be forgotten) or to carry cards or keys (which can be
stolen). Among all biometric techniques, iris recognition has been
traditionally regarded as one of the most reliable and accurate
biometric identification system available \cite{[Jain99Kluwer]}.
Additionally, the iris is highly stable over a person's lifetime and
lends itself to noninvasive identification because it is an
externally visible internal organ \cite{[Monro07]}.

However, in spite of these advantages, biometric systems have some
drawbacks \cite{[schneier99usesAbuses]}: $i)$ the lack of secrecy
(e.g. everybody knows our face or could get our fingerprints), and
$ii)$ the fact that a biometric trait can not be replaced (if we
forget a password we can easily generate a new one, but no new
fingerprint can be generated if an impostor ``steals'' it).
Moreover, biometric systems are vulnerable to external attacks which
could decrease their level of security. In \cite{[Ratha01]} Ratha
\emph{et al}. identified and classified eight possible attack points
to biometric recognition systems. These vulnerability points,
depicted in Figure~\ref{fig:overall-system}, can broadly be divided
into two main groups:

\begin{figure}[htb]
     \centering
     \includegraphics[width=.8\linewidth]{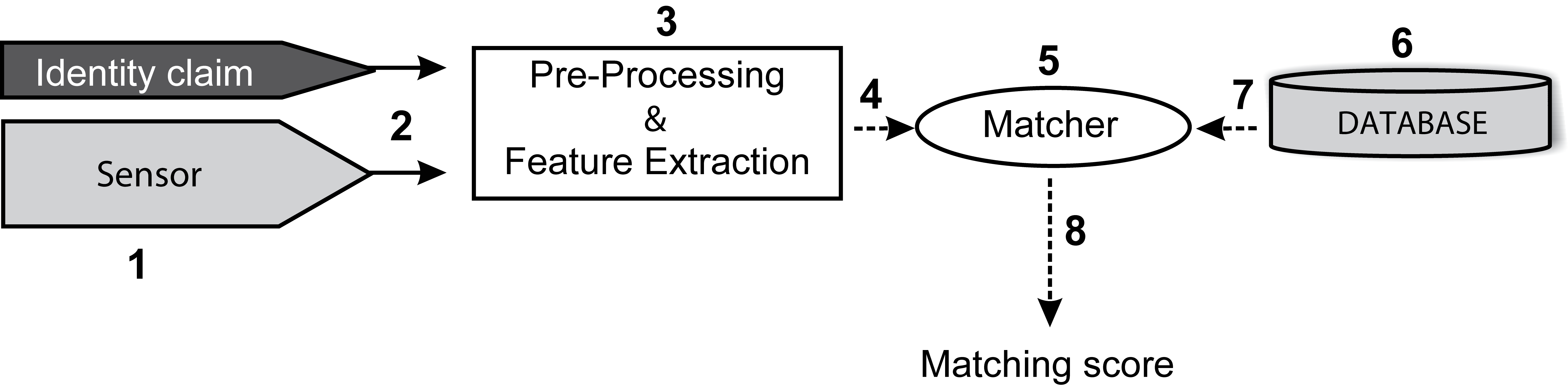}
     \caption{Architecture of an automated biometric verification system.
     Possible attack points are numbered from 1 to 8.}
     \label{fig:overall-system}
\end{figure}

\begin{itemize}
    \item \textbf{Direct attacks}. Here, the sensor is attacked
    using synthetic biometric samples, e.g. gummy fingers
    (point 1 in Figure~\ref{fig:overall-system}).
    It is worth noting that in this type of attacks no specific
    knowledge about the system is needed.
    Furthermore, the attack is carried out in the analog domain, outside
    the digital limits of the system, so digital protection
    mechanisms (digital signature, watermarking, etc) cannot be
    used.
    \\

    \item \textbf{Indirect attacks}. This group includes all the remaining
    seven points of attack identified in Figure~\ref{fig:overall-system}.
    Attacks 3 and 5 might be carried out using a Trojan Horse that
    bypasses the system modules. In attack 6, the system database is
    manipulated. The remaining points of attack (2, 4, 7 and 8)
    exploit possible weak points in the communication
    channels of the system. In opposition to direct attacks, in this
    case the intruder needs to have some additional information about
    the internal working of the system and, in most cases,
    physical access to some of the application components.
    Most of the works reporting indirect attacks use some
    type of variant of the hill climbing technique introduced in
    \cite{[Soutar99]}.

\end{itemize}

In this work we concentrate our efforts in studying direct attacks
on iris-based verification systems. For this purpose we have built a
database with synthetic iris images generated from 27 users of the
BioSec multi-modal baseline corpus \cite{[Fierrez07]}. This paper is
structured as follows. In Sect.~\ref{sec:fakedatabase} we detail the
process followed for the creation of the fake iris, and the database
used in the experiments is presented. The experimental protocol,
some results and further discussion are reported in
Sect.~\ref{sec:experiments}. Conclusions are finally drawn in
Sect.~\ref{sec:conclusion}.

\begin{figure}[htb]
\centering
    \begin{minipage}[b]{0.75\linewidth}
        \centering
        \centerline{\epsfig{figure=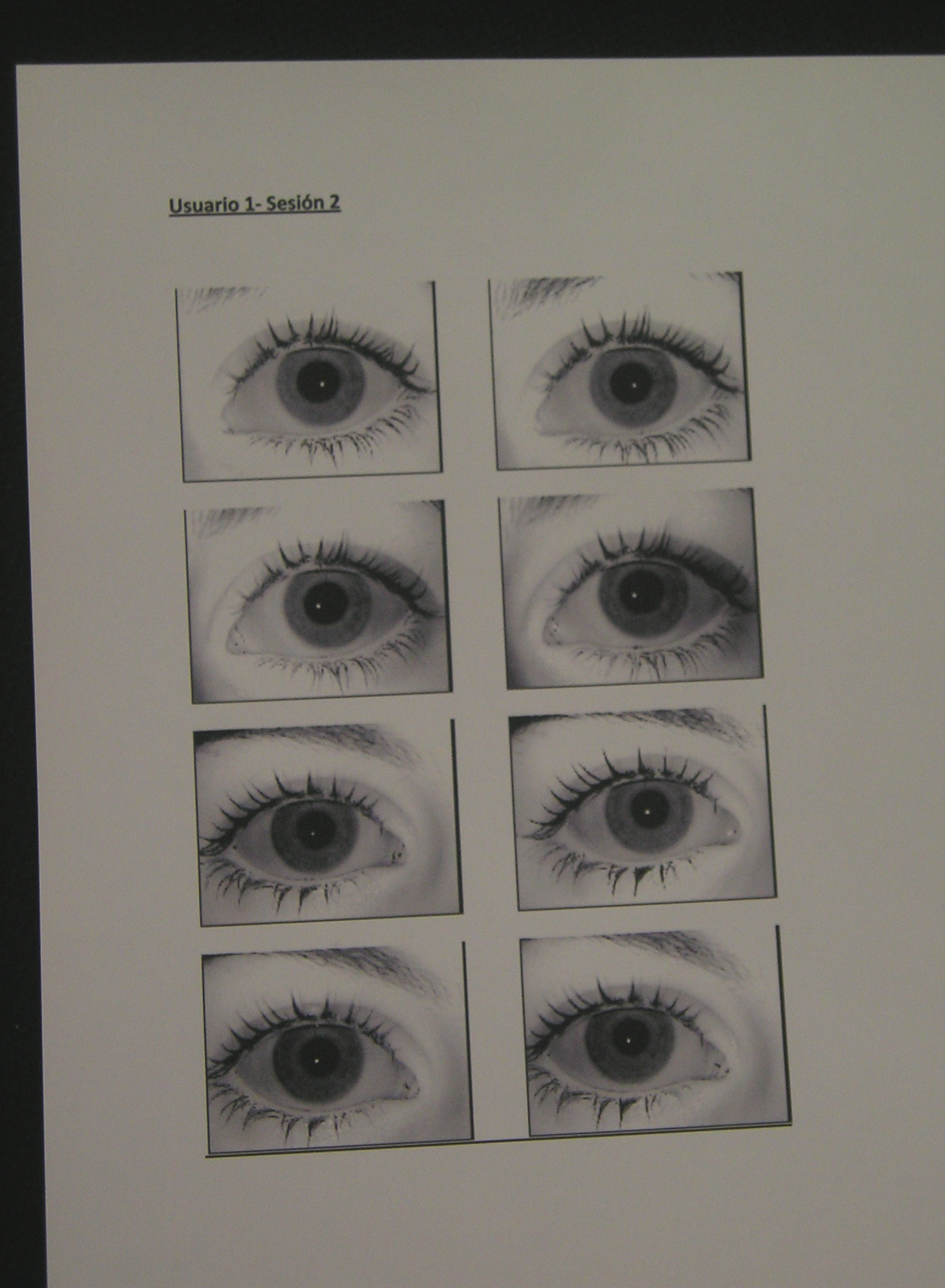,width=0.8\linewidth}}
    \end{minipage}
    \caption{Iris capture preparation.}
    \label{fig:folio_iris}
\end{figure}

\begin{figure}[h]
\centering
    \begin{minipage}[b]{0.75\linewidth}
        \centering
        \centerline{\epsfig{figure=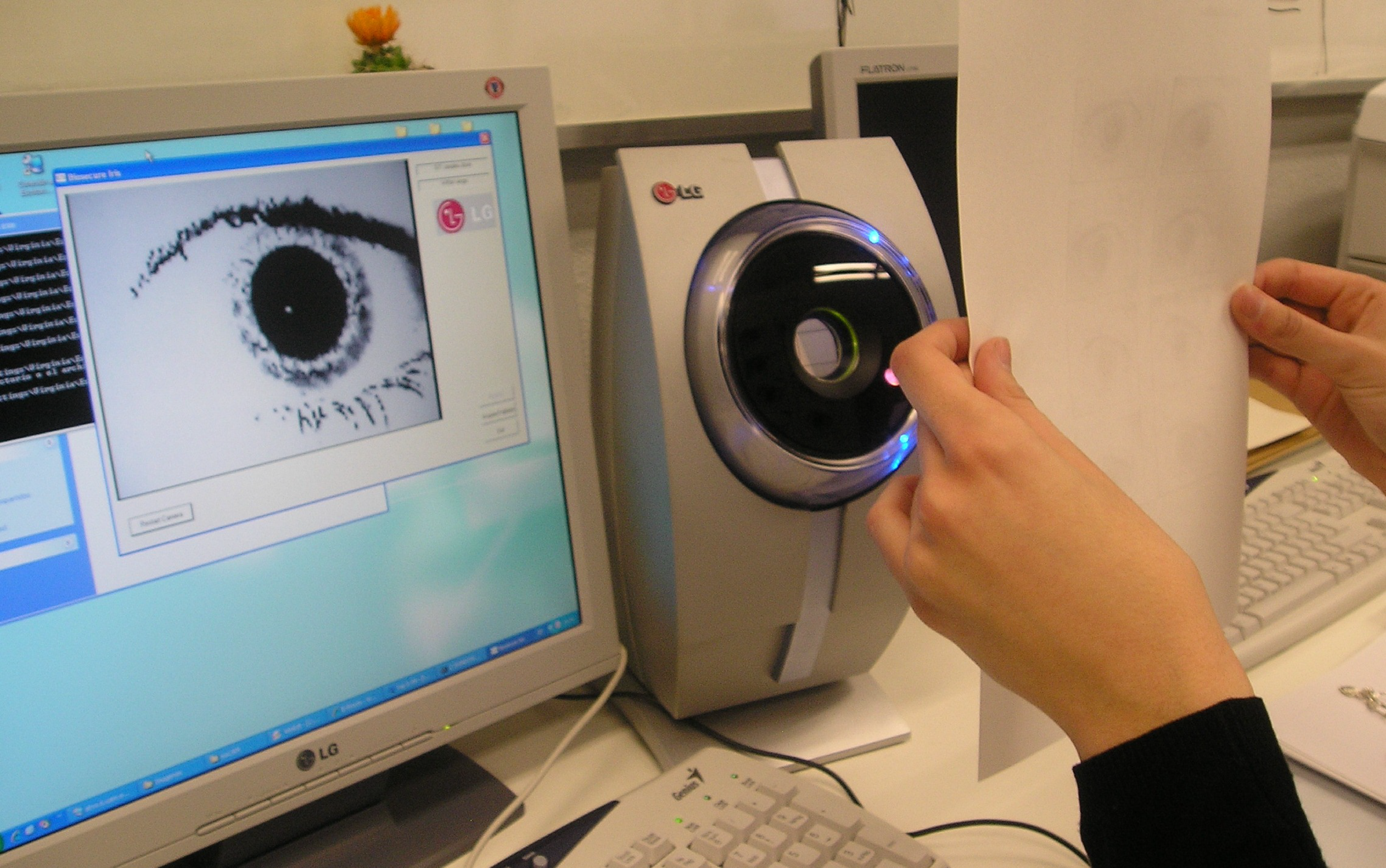,width=0.8\linewidth}}
    \end{minipage}
    \caption{Capturing fake iris.}
    \label{fig:captura}
\end{figure}

\begin{table}
    \centering
    \scriptsize
    \begin{tabular}{|c|c|c|}
        \hline
        \hspace{2mm} PRINTER \hspace{2mm} & PAPER & \hspace{2mm} PREPROCESSING \cite{[Gonzalez]} \hspace{2mm} \\
        \hline
        Ink Jet &   White paper           & Histogram equalization \\
        Laser   &   Recycled paper        & Noise filtering         \\
                 &  Photographic paper    & Open/close         \\
                 &  \hspace{2mm} High resolution paper \hspace{2mm} & Top hat         \\
                 &  Butter paper          &          \\
                 &  Cardboard             &          \\
        \hline
        \multicolumn{3}{c}{} \\
    \end{tabular}
    \caption{Options tested for fake iris generation.}
    \label{tab:quality variables}
\end{table}

\section{Fake Iris Database}
\label{sec:fakedatabase}

A new iris database has been created using iris images from 27 users
of the BioSec baseline database \cite{[Fierrez07]}. The process is
divided into three steps: $i)$ first original images are
preprocessed for a better afterwards quality, then $ii)$ they are
printed on a piece of paper using a commercial printer as shown in
Figure~\ref{fig:folio_iris}, and lastly, $iii)$ printed images are
presented at the iris sensor, as can be seen in
Figure~\ref{fig:captura}, obtaining the fake image.


\subsection{Fake iris generation method}

To correctly create a new database, it is necessary to take into
account factors affecting the quality of acquired fake images. The
main variables with significant importance for iris quality are
found to be: preprocessing of original images, printer type and
paper type.

\begin{figure}[htb]
    \centering
    \begin{minipage}[b]{0.35\linewidth}
        \centering
            \epsfig{figure=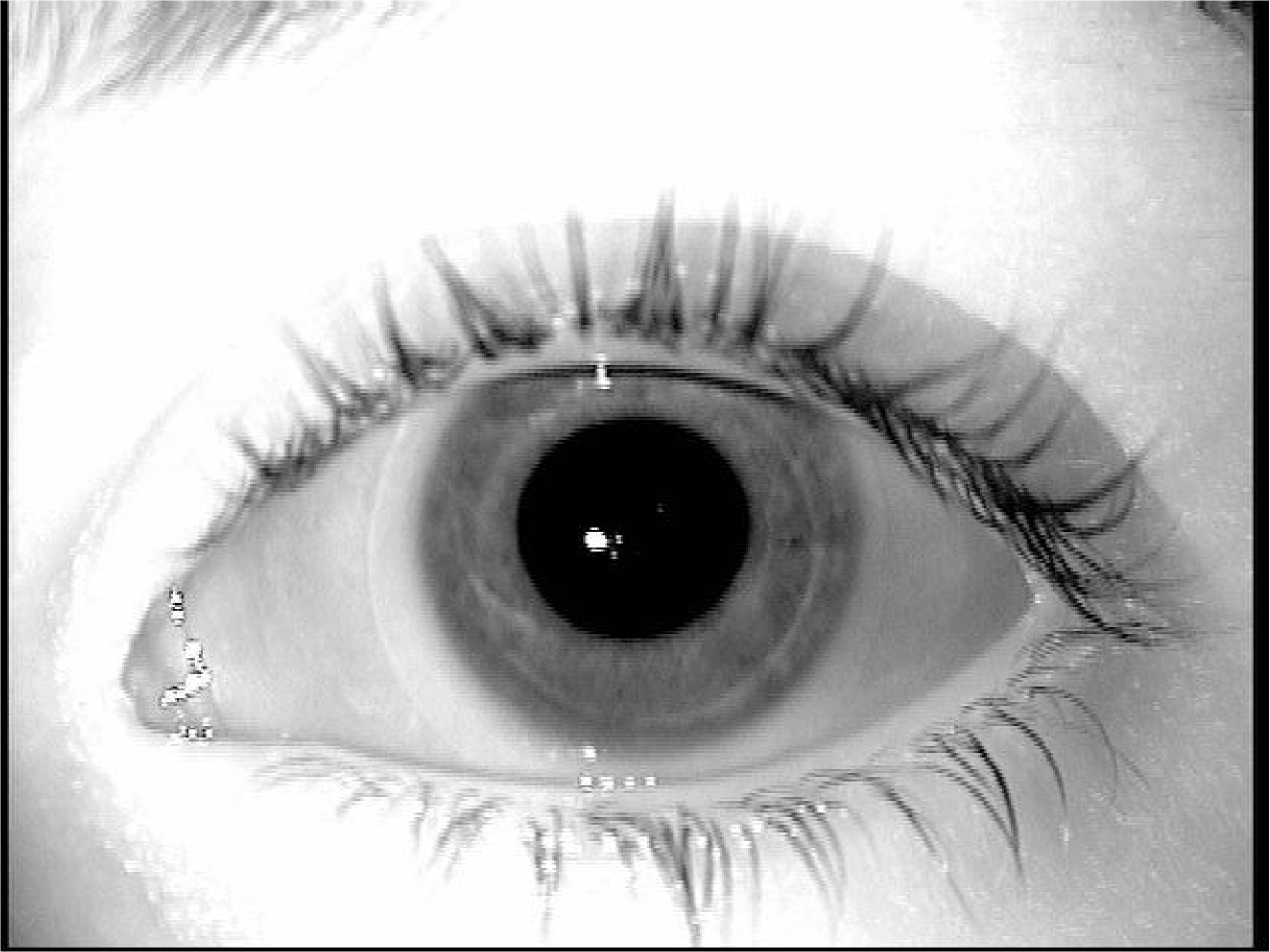,width=1.0\linewidth}
            \centerline{\scriptsize (a) Original image - no enhancement}\medskip
    \end{minipage}
    \begin{minipage}[b]{0.1\linewidth}
        \centering
            \epsfig{figure=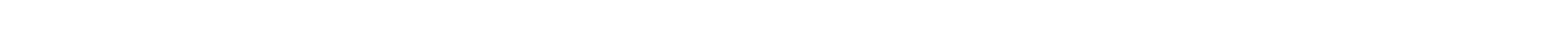,width=1.0\linewidth}
    \end{minipage}
    \begin{minipage}[b]{0.35\linewidth}
        \centering
        \epsfig{figure=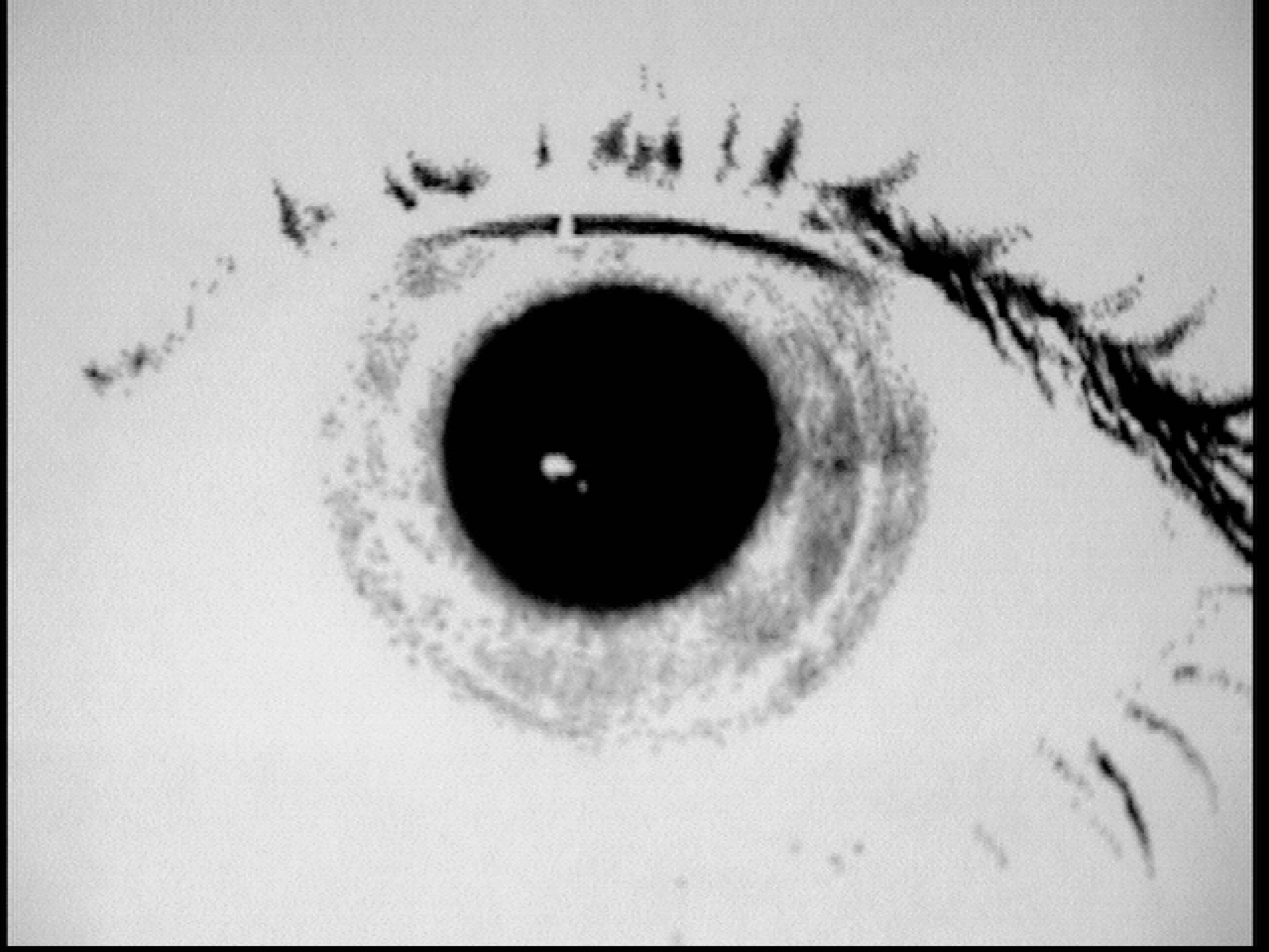,width=1.0\linewidth}
        \centerline{\scriptsize (b) Fake image - no enhancement}\medskip
    \end{minipage}
    \begin{minipage}[b]{0.35\linewidth}
        \centering
            \epsfig{figure=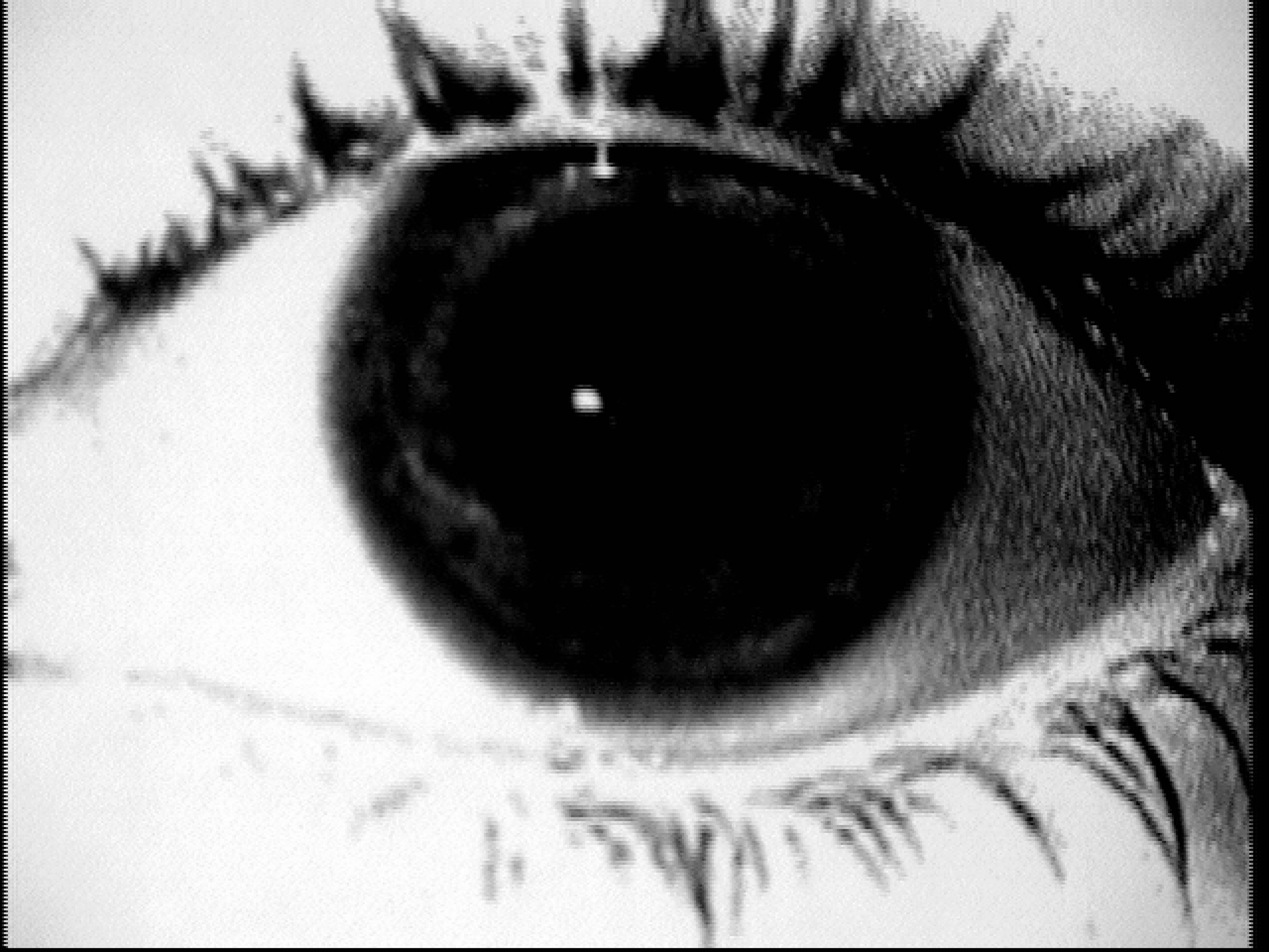,width=1.0\linewidth}
            \centerline{\scriptsize (c) Fake image - histogram equalization}\medskip
    \end{minipage}
    \begin{minipage}[b]{0.1\linewidth}
        \centering
            \epsfig{figure=BLANK.png,width=1.0\linewidth}
    \end{minipage}
    \begin{minipage}[b]{0.35\linewidth}
        \centering
            \epsfig{figure=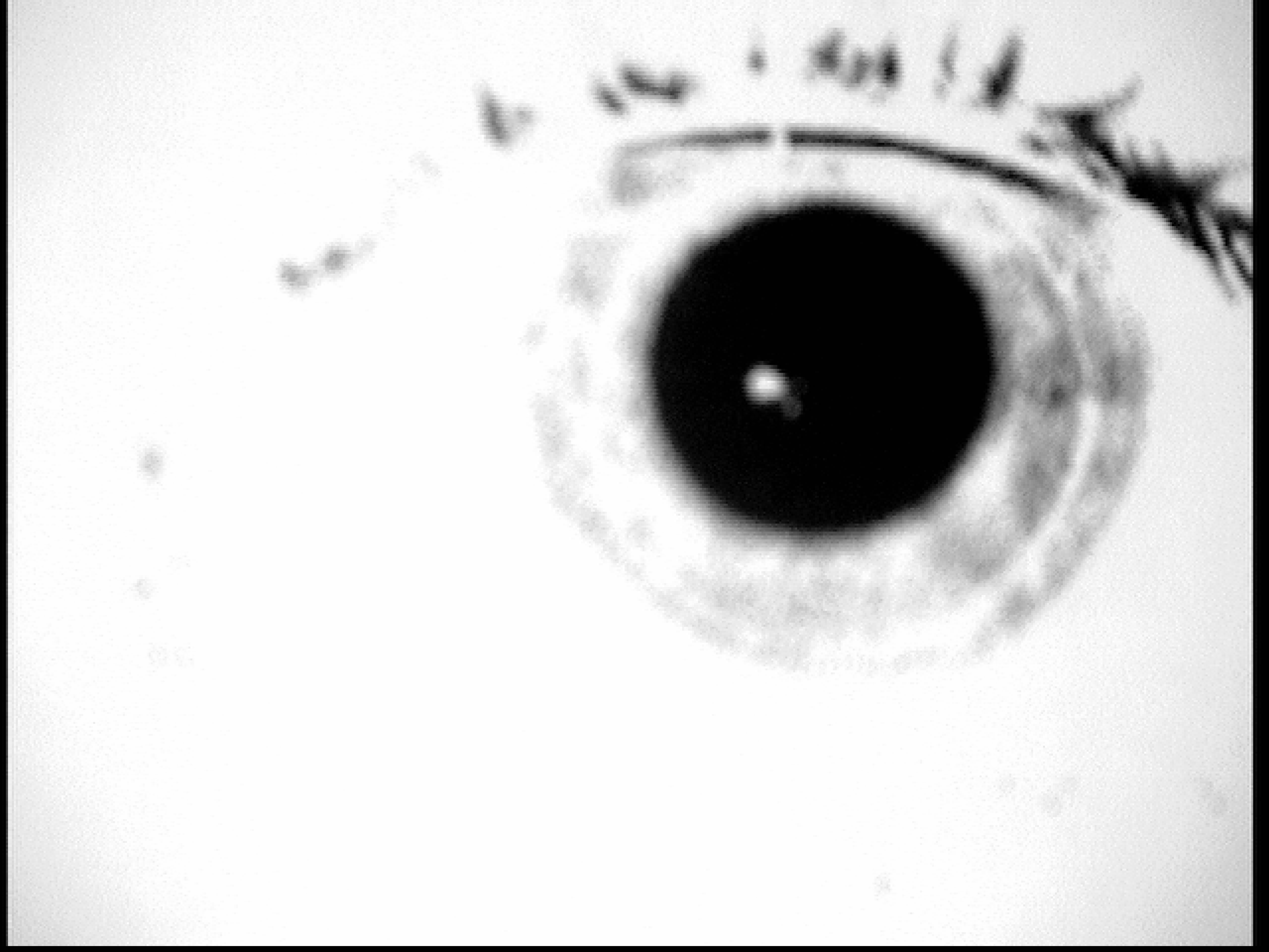,width=1.0\linewidth}
            \centerline{\scriptsize (d) Fake image - noise filtering}\medskip
    \end{minipage}
    \begin{minipage}[b]{0.35\linewidth}
        \centering
            \epsfig{figure=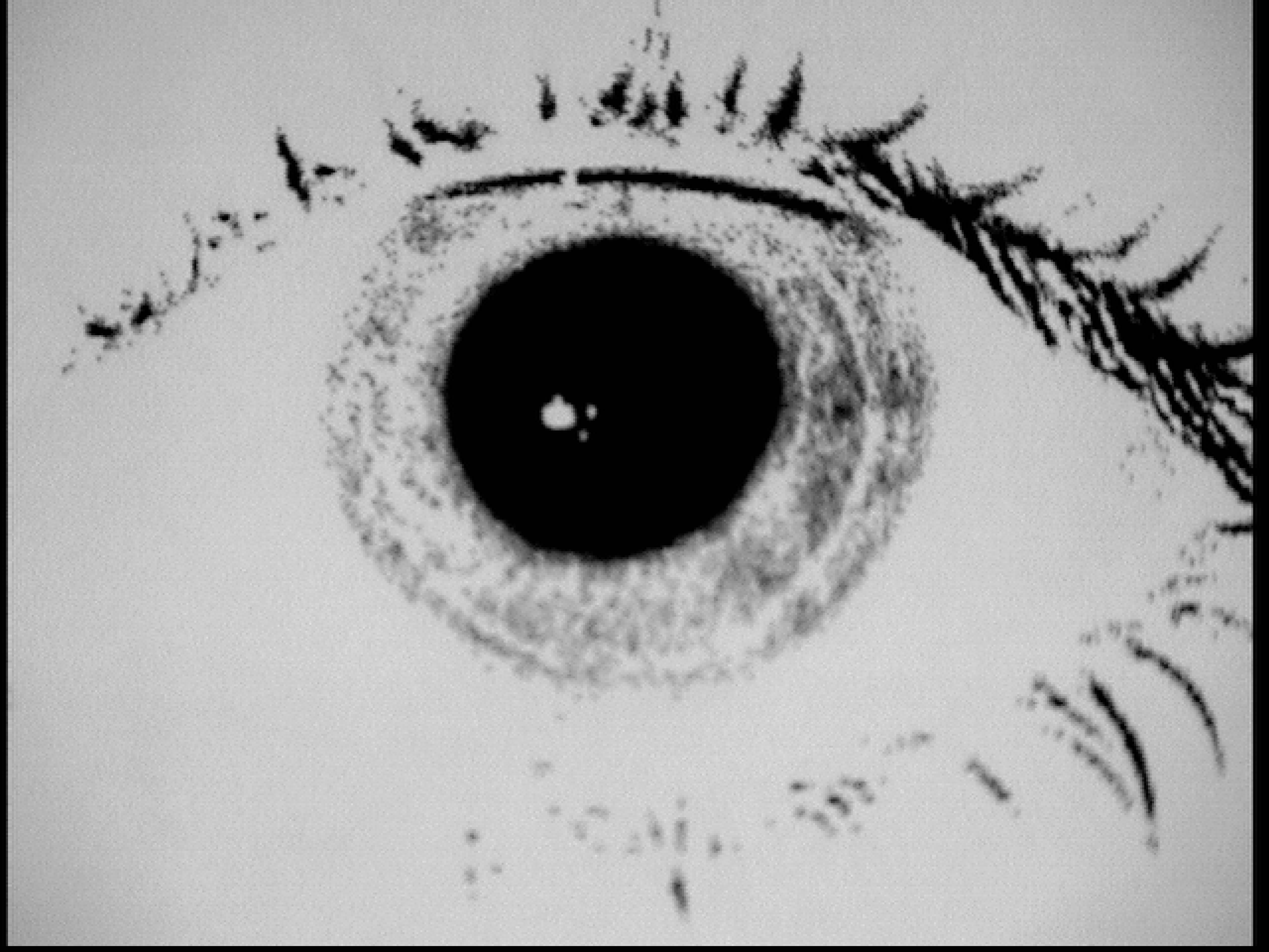,width=1.0\linewidth}
            \centerline{\scriptsize (e) Fake image - TopHat}\medskip
    \end{minipage}
    \begin{minipage}[b]{0.1\linewidth}
        \centering
            \epsfig{figure=BLANK.png,width=1.0\linewidth}
    \end{minipage}
    \begin{minipage}[b]{0.35\linewidth}
        \centering
            \epsfig{figure=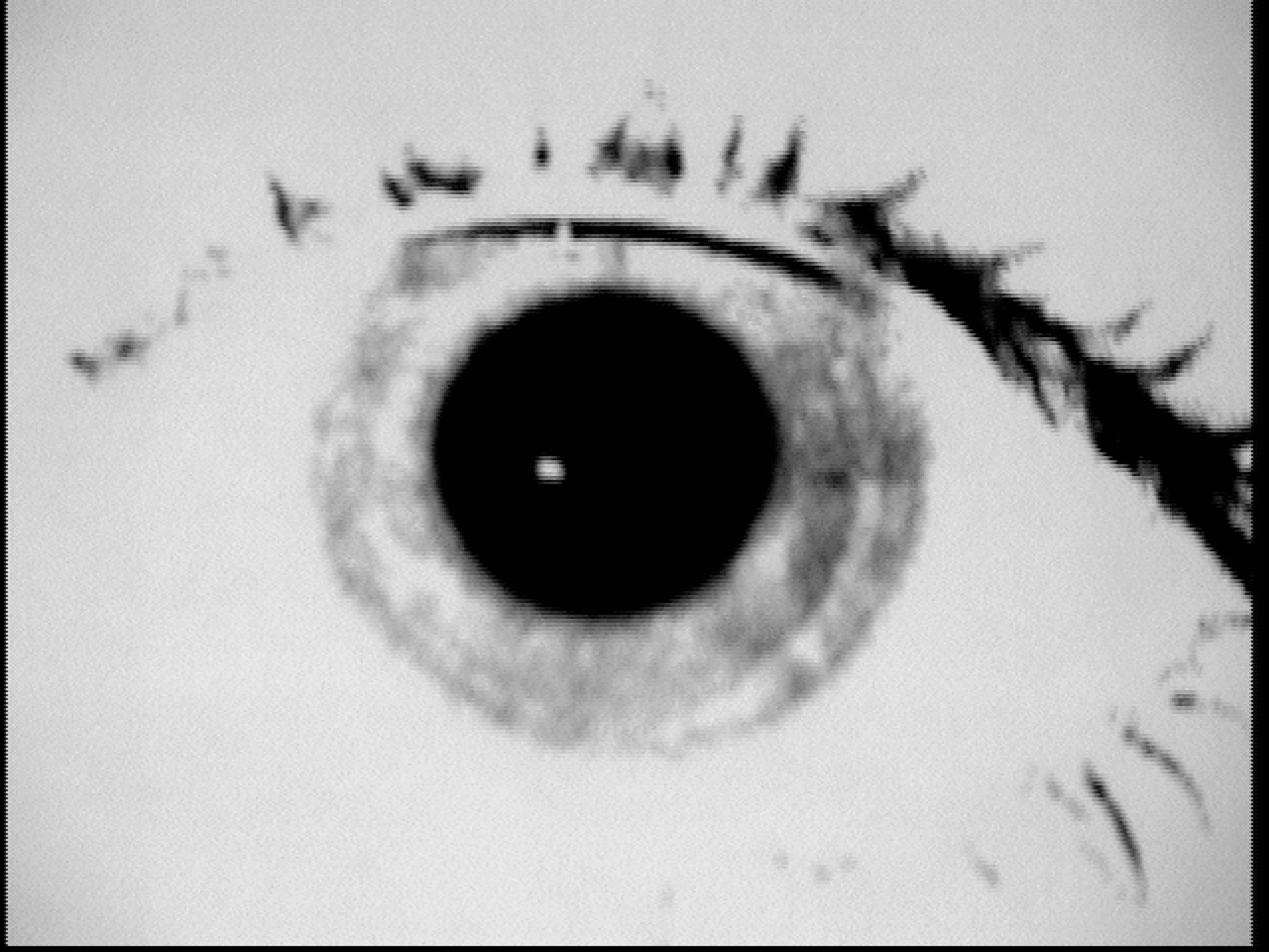,width=1.0\linewidth}
            \centerline{\scriptsize (f) Fake image - Open+TopHat}\medskip
    \end{minipage}
    \caption{Acquired fake images with different modifications using high quality paper and inkjet printer.}
    \label{fig:HighImag}
\end{figure}

We tested two different printers: a HP Deskjet 970cxi (inkjet
printer) and a HP LaserJet 4200L (laser printer). They both give
fairly good quality. On the other hand, we observed that the quality
of acquired fake images depends on the type of paper used. Here
comes the biggest range of options. All the tested types appear in
Table~\ref{tab:quality variables}. In our experiments, the
preprocessing is specially important since it has been observed that
the iris camera does not capture correctly original images printed
without previous modifications. Therefore we have tested different
enhancement methods before printing in order to acquire good quality
fake images. The options tested are also summarized in
Table~\ref{tab:quality variables}. By analyzing all the
possibilities with a few images, the combination that gives the best
segmentation results and therefore the best quality for the
afterwards comparison has been found to be the inkjet printer, with
high resolution paper and an Open-TopHat preprocessing step. In
Figure~\ref{fig:HighImag}, examples using different preprocessing
techniques with this kind of paper and inkjet printer are shown.

\subsection{Database}

The fake iris database follows the same structure of the original
BioSec database. Therefore, data for the experiments consists of 27
users $\times$ 2 eyes $\times$ 4 images $\times$ 2 sessions = 432
fake iris images, and its corresponding real images. Acquisition of
fake images has been carried out with the same iris camera used in
BioSec, a LG IrisAccess EOU3000.

\section{Experiments}
\label{sec:experiments}

\subsection{Recognition system} \label{sec:system}

We have used for our experiments the iris recognition
system\footnote{The source code can be freely downloaded from
\url{www.csse.uwa.edu.au/~pk/studentprojects/libor/sourcecode.html}}
developed by Libor Masek \cite{systemLiborB}. It consists of the
following sequence of steps that are described next: segmentation,
normalization, encoding and matching.

\begin{figure}[htb]
 \centering
    \begin{minipage}[b]{0.85\linewidth}
        \centering
        \centerline{
            \epsfig{figure=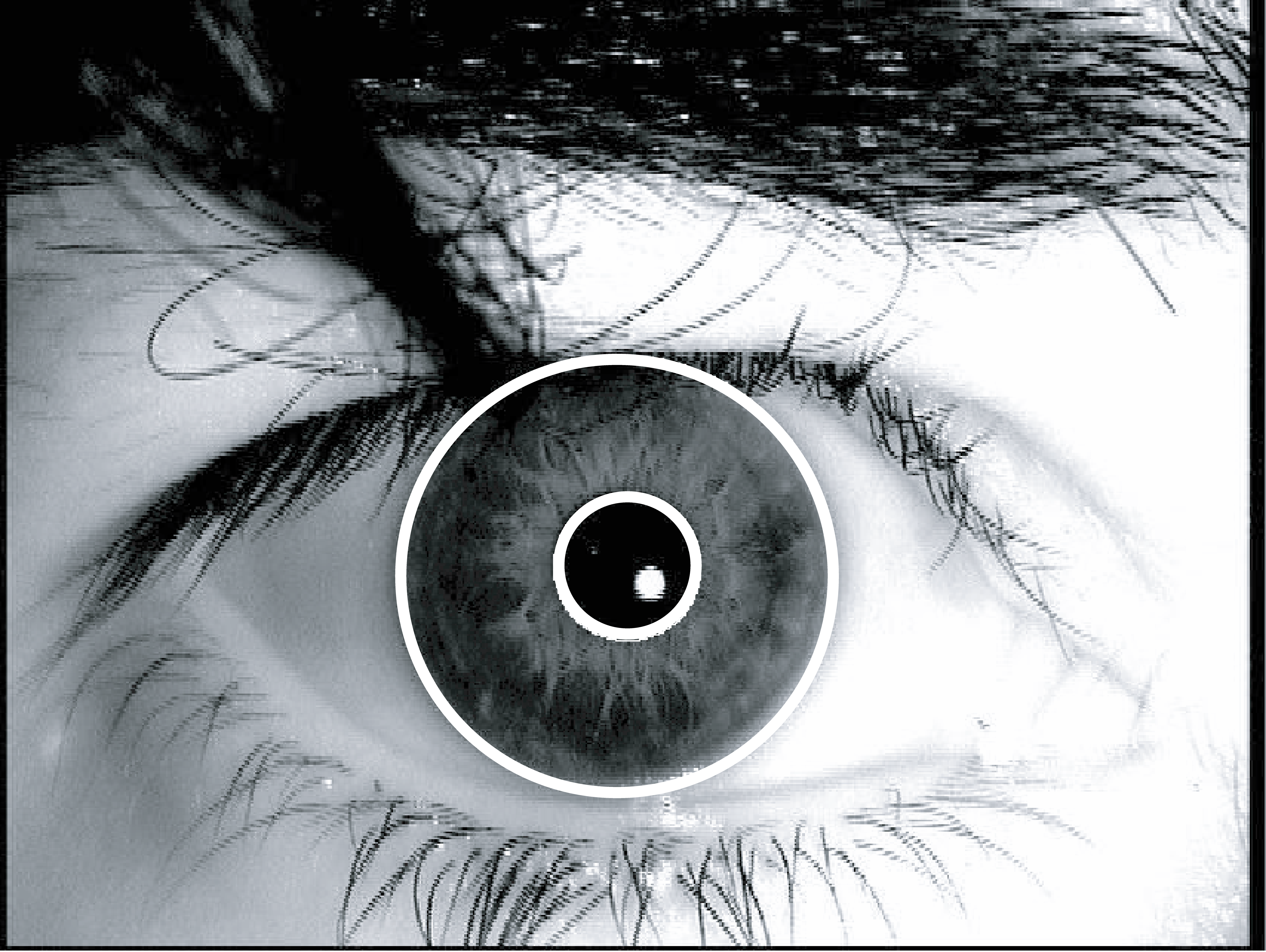,width=0.4\linewidth}
            \epsfig{figure=BLANK.png,width=0.1\linewidth}
            \epsfig{figure=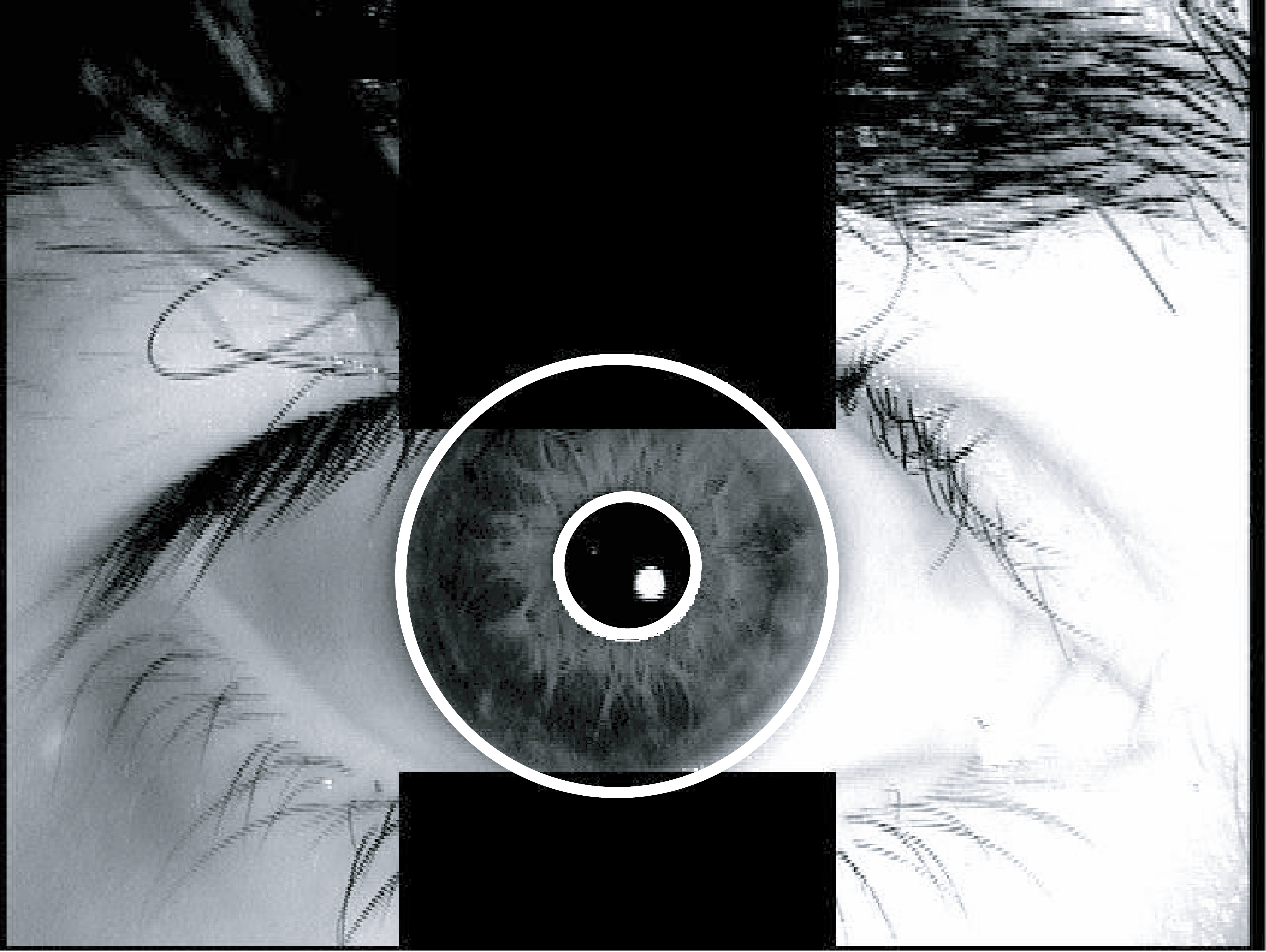,width=0.4\linewidth}
        }
        \centerline{\scriptsize (a) Original image and noise image}\medskip
    \end{minipage}
    \begin{minipage}[b]{0.9\linewidth}
        \centering
        \centerline{
            \epsfig{figure=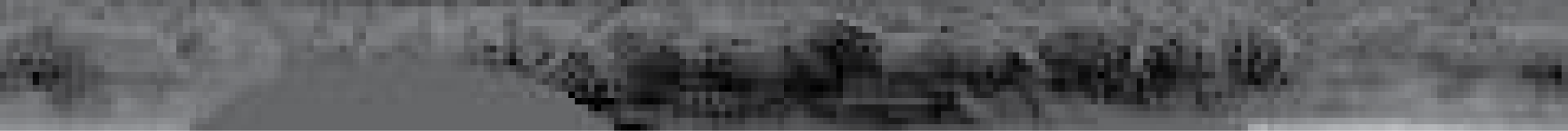,width=0.9\linewidth}
        }
        \centerline{\scriptsize (b) Normalized iris pattern}\medskip
    \end{minipage}
    \begin{minipage}[b]{0.9\linewidth}
        \centering
        \centerline{
            \epsfig{figure=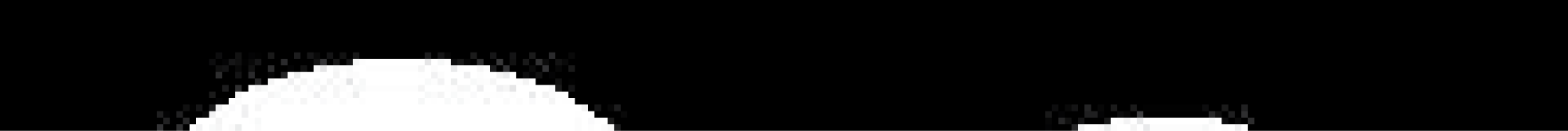,width=0.9\linewidth}
        }
        \centerline{\scriptsize (c) Noise mask}\medskip
    \end{minipage}
    \caption{Examples of the normalization step.}
    \label{fig:normalization}
\end{figure}

For iris segmentation, the system uses a circular Hough transform in
order to detect the iris and pupil boundaries. Iris boundaries are
modeled as two circles. The system also performs an eyelids removal
step. Eyelids are isolated first by fitting a line to the upper and
lower eyelid using a linear Hough transform (see
Figure~\ref{fig:normalization}(a) right, in which the eyelid lines
correspond to the border of the black blocks). Eyelashes detection
by histogram thresholding is available in the source code, but it is
not performed in our experiments. Although eyelashes are quite dark
compared with the surrounding iris region, other iris areas are
equally dark due to the imaging conditions. Therefore, thresholding
to isolate eyelashes would also remove important iris regions.
However, eyelash occlusion has been found to be not very prominent
in our database.

Normalization of iris regions is performed using a technique based
on Daugman's rubber sheet model \cite{[Daugman04]}. The center of
the pupil is considered as the reference point, based on which a 2D
array is generated consisting of an angular-radial mapping of the
segmented iris region. In Figure~\ref{fig:normalization}, an example
of the normalization step is depicted.

Feature encoding is implemented by convolving the normalized iris
pattern with 1D Log-Gabor wavelets. The rows of the 2D normalized
pattern are taken as the 1D signal, each row corresponding to a
circular ring on the iris region. It uses the angular direction
since maximum independence occurs in this direction. The filtered
output is then phase quantized to four levels using the Daugman
method \cite{[Daugman04]}, with each filtering producing two bits of
data. The output of phase quantization is a grey code, so that when
going from one quadrant to another, only 1 bit changes. This will
minimize the number of bits disagreeing, if say two intra-class
patterns are slightly misaligned and thus will provide more accurate
recognition \cite{systemLiborB}. The encoding process produces a
binary template and a corresponding noise mask which represents the
eyelids areas (see Figure~\ref{fig:normalization} (c)).

For matching, the Hamming distance is chosen as a metric for
recognition. The Hamming distance employed incorporates the noise
mask, so that only significant bits are used in calculating the
Hamming distance between two iris templates. The modified Hamming
distance formula is given by

$$
HD=\frac{1}{N-\sum_{k=1}^{N}Xn_k(OR)Yn_k}\cdot\sum_{j=1}^{N}X_j(XOR)Y_j(AND)Xn'_j(AND)Yn'_j
$$

where $X_j$ and $Y_j$ are the two bitwise templates to compare,
$Xn_j$ and $Yn_j$ are the corresponding noise masks for $X_j$ and
$Y_j$, and $N$ is the number of bits represented by each template.

In order to account for rotational inconsistencies, when the Hamming
distance of two templates is calculated, one template is shifted
left and right bitwise and a number of Hamming distance values are
calculated from successive shifts \cite{[Daugman04]}. This method
corrects for misalignments in the normalized iris pattern caused by
rotational differences during imaging. From the calculated distance
values, the lowest one is taken.

\subsection{Experimental Protocol}

For the experiments, each eye in the database is considered as a
different user. In this way, we have two sessions with 4 images each
for 54 users (27 donors $\times$ 2 eyes per donor).

Two different attack scenarios are considered in the
experiments and compared to the system normal operation mode:

\begin{itemize}
\item \textbf{Normal Operation Mode (NOM)}: both the enrollment and the test are carried out
with a real iris. This is used as the reference scenario. In
this context the FAR (False Acceptance Rate) of the system is
defined as the number of times an impostor using his own iris
gains access to the system as a genuine user, which can be
understood as the robustness of the system against a zero-effort
attack. The same way, the FRR (False Rejection Rate) denotes the
number of times a genuine user is rejected by the system.

\item \textbf{Attack 1}: both the enrollment and the test are carried out with
a fake iris. In this case the attacker enrolls to the system
with the fake iris of a genuine user and then tries to
access the application also with a fake iris of the same user. In this
scenario an attack is unsuccessful (i.e. the system repels the
attack) when the impostor is not able to
access the system using the fake iris. Thus, the attack
success rate (SR) in this scenario can be computed as:
$\textrm{SR}=1-\textrm{FRR}$.

\item \textbf{Attack 2}: the enrollment is performed using a real iris, and
tests are carried out with fake iris. In this case the
genuine user enrolls with his/her iris and the attacker tries to
access the application with the fake iris of the legal user.
A successful attack is accomplished when the system confuses a fake
iris with its corresponding genuine iris, i.e.,
$\textrm{SR}=\textrm{FAR}$.
\end{itemize}

\begin{figure*}[htb]
    \centering
    \begin{minipage}[b]{0.35\linewidth}
        \centering
            \epsfig{figure=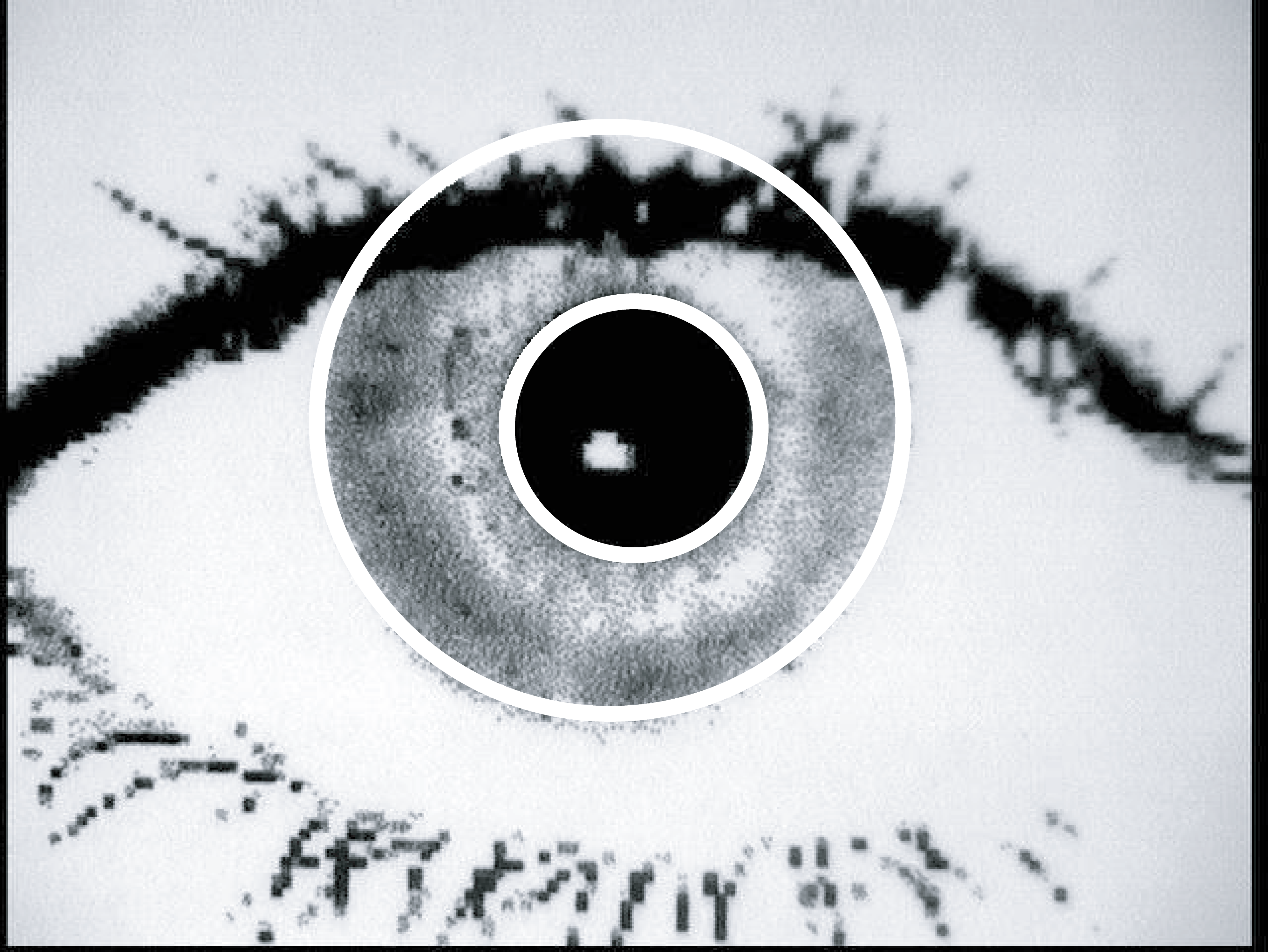,width=1.0\linewidth}
            \epsfig{figure=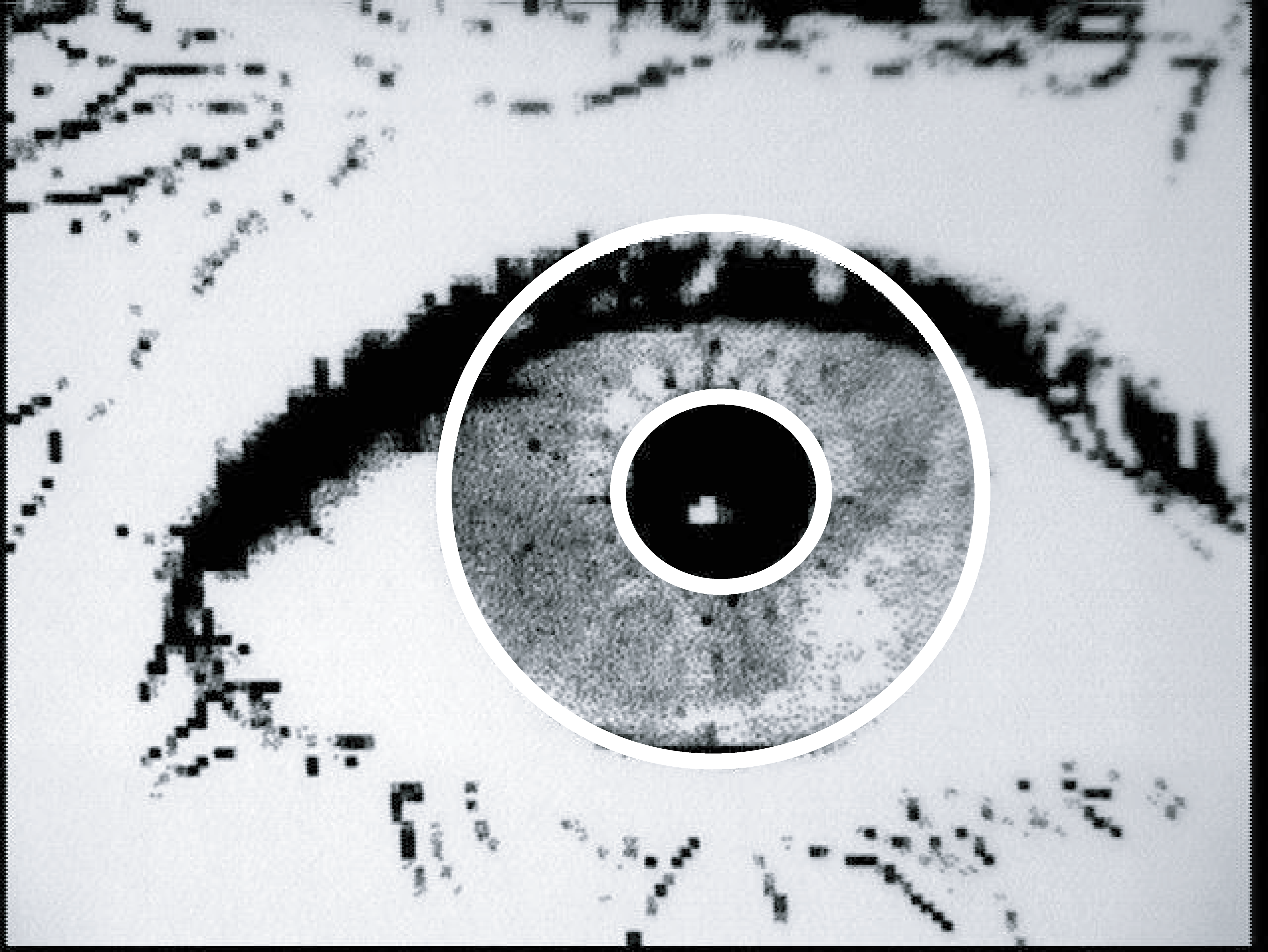,width=1.0\linewidth}
            \centerline{\scriptsize (a) Correct iris detection }\medskip
    \end{minipage}
    \begin{minipage}[b]{0.1\linewidth}
        \centering
            \epsfig{figure=BLANK.png,width=1.0\linewidth}
    \end{minipage}
    \begin{minipage}[b]{0.35\linewidth}
        \centering
        \epsfig{figure=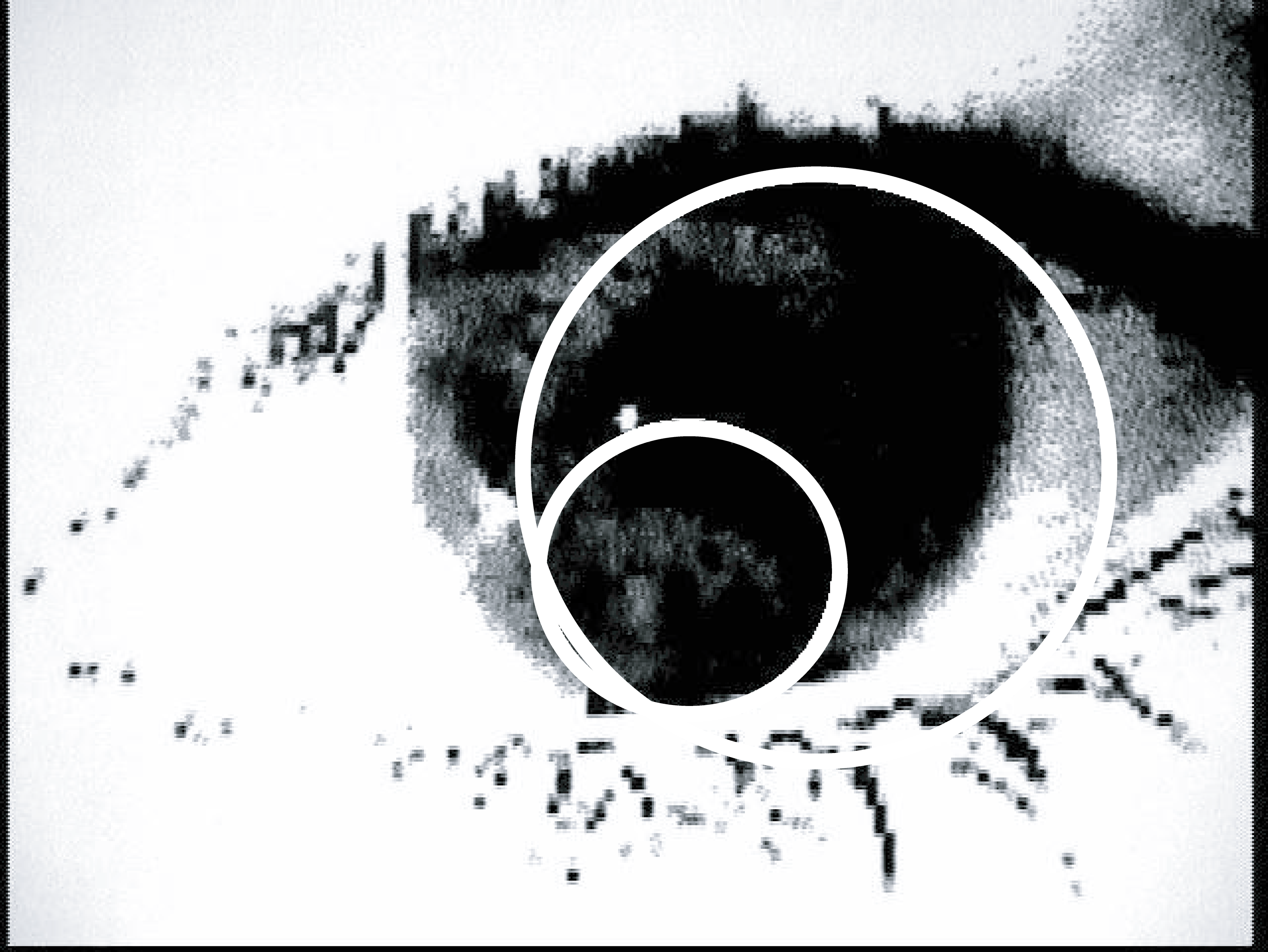,width=1.0\linewidth}
        \epsfig{figure=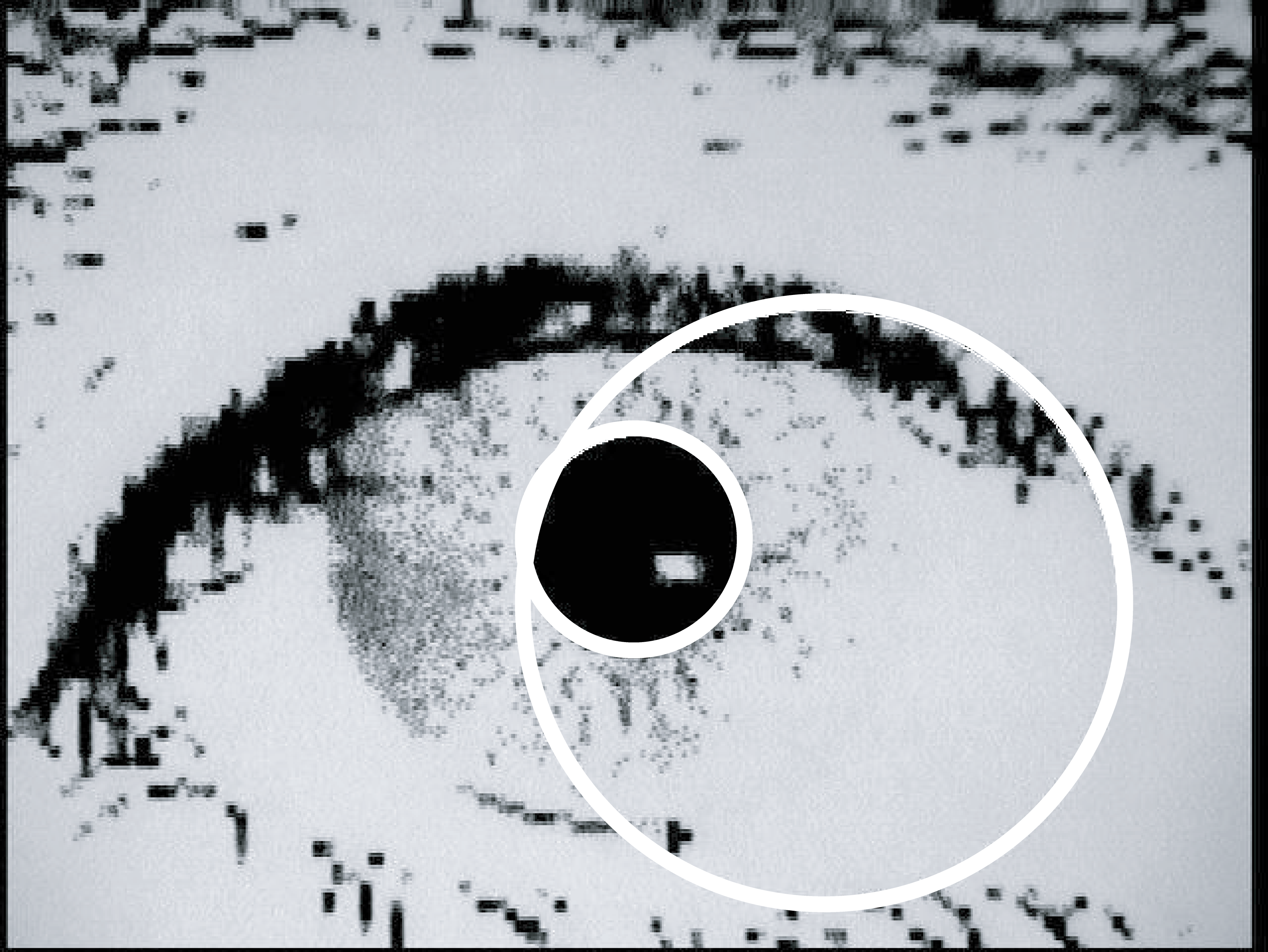,width=1.0\linewidth}
        \centerline{\scriptsize (b) Incorrect iris detection}\medskip
    \end{minipage}
    \caption{Examples of fake images with correct iris detection (left) and incorrect iris detection (right).}
    \label{fig:imagenes_results}
\end{figure*}

In order to compute the performance of the system in the normal
operation mode, the experimental protocol is as follows.
For a given user, all the images of the first session
are considered as enrolment templates. Genuine matchings are
obtained by comparing the templates to the corresponding images of
the second session from the same user. Impostor
matchings are obtained by comparing one randomly selected template
of a user to a randomly selected iris image of the second session
from the remaining users. Similarly, to compute the FRR in attack 1, all
the fake images of the first session of each user are compared with the corresponding
fake images of the second session. In the attack 2 scenario, only the impostor scores are computed matching all the 4
original samples of each user with its 4 fake samples of the second session. In our experiments, not all the images
were segmented successfully by the recognition system. As a result, it was not possible to use all the
eye images for testing experiments.

\subsection{Results}

The number of correctly segmented images were 348 for the original
database (80.56\% of the 432 available) and 166 for the fake
database (38.43\% of the 432). In Figure~\ref{fig:imagenes_results},
several examples of fake images with correct and incorrect iris
detection are plotted. The rate of correctly segmented images for
the original database is consistent with that reported in the
description of the recognition system used in this paper, with which
a segmentation rate of around 83\% is attained on the CASIA database
\cite{systemLiborB}. Regarding fake images, it is worth noting than
nearly 40\% of them pass through the segmentation and normalization
stages, and they are input into the feature extraction and matching
stages. It should be noted that the version of the CASIA database
used in \cite{systemLiborB} provided good segmentation, since pupil
regions of all iris images were automatically detected and replaced
with a circular region of constant intensity to mask out the
specular reflections, thus making iris boundaries clearly
distinguishable.

In Table~\ref{tab:results} we show the Success Rate (SR) of the
direct attacks against the recognition system at four different
operating points, considering only the matchings between correctly
segmented images. The decision threshold is fixed to reach a
FAR=\{0.1, 1, 2, 5\} $\%$ in the normal operation mode (NOM), and
then the success rate of the two proposed attacks is computed. We
observe that in all the operating points, the system is highly
vulnerable to the two attacks (i.e. a success rate of 50\% or higher
is observed). This is specially evident as the FAR in the normal
operation mode is increased. Also, higher success rates are observed
for attack 1. For this kind of attack, an intruder would be
correctly enrolled in the system using a fake image of another
person and at a later date, he/she would be granted access to the
system also using a fake image.

\begin{table}
    \centering
    \small
    \begin{tabular}{|c|c|c|}

        \hline
        \textbf{NOM} & \hspace{2mm} \textbf{Attack 1} \hspace{2mm} & \hspace{2mm} \textbf{Attack 2} \hspace{2mm} \\ \hline

        \hspace{2mm} FAR - FRR (\%) \hspace{2mm}  & SR (\%) & SR (\%) \\ \hline \hline

        0.1 - 12.71  & 57.41 & 49.32  \\\hline
        1 - 8.70  & 74.07 & 66.06  \\\hline
        2 - 7.86  & 76.85 & 68.78  \\\hline
        5 - 6.19  & 82.41 & 73.30  \\\hline
        \multicolumn{3}{l}{}   \\
    \end{tabular}
    \caption{Evaluation of the verification system to direct attacks. NOM refers to the system normal operation mode and
    SR to the success rate of the attack.}
    \label{tab:results}
\end{table}

\section{Conclusion}
\label{sec:conclusion}

An evaluation of the vulnerabilities to direct attacks of iris-based
verification systems has been presented. The attacks have been
evaluated using fake iris images created from real iris of the
BioSec baseline database. We printed iris images with a commercial
printer and then, we presented the images to the iris sensor.
Different factors affecting the quality of acquired fake images have
been studied, including preprocessing of original images, printer
type and paper type. We have chosen the combination giving the best
quality and then, we have built a database of fake images from 54
eyes, with 8 iris images per eye. Acquisition of fake images has
been carried out with the same iris camera used in BioSec.

Two attack scenarios have been compared to the normal operation mode
of the system using a publicly available iris recognition system.
The first attack scenario considers enrolling to the system and
accessing it with fake iris. The second one represents accessing a
genuine account with fake iris. Results showed that the system is
highly vulnerable to the two evaluated attacks. We also observed
that about 40\% of the fake images were correctly segmented by the
system. When that this happens, the intruder is granted access with
high probability, being the success rate of the two attacks of 50\%
or higher.

Liveness detection procedures are possible countermeasures against
direct attacks. For the case of iris recognition systems, light
reflections or behavioral features like eye movement, pupil response
to a sudden lighting event, etc. have been proposed
\cite{[DaugmanOnlineIrisLiveness],[Pacut06]}. This research
direction will be the source of future work. We will also explore
the use of another type of iris sensors, as the OKI's hand-held iris
sensor used in the CASIA
database\footnote{http://www.cbsr.ia.ac.cn/databases.htm}.

\small

\subsubsection*{Acknowledgments.}

This work has been supported by Spanish project
TEC2006-13141-C03-03, and by European Commission IST-2002-507634
Biosecure NoE. Author F. A.-F. is supported by a FPI Fellowship from
Consejeria de Educacion de la Comunidad de Madrid. Author J. G. is
supported by a FPU Fellowship from the Spanish MEC. Author J. F. is
supported by a Marie Curie Fellowship from the European Commission.

\bibliographystyle{splncs}

\begin{thebibliography}{}

\end{thebibliography}


\begin{thebibliography}{10}

\bibitem{[Jain06]}
Jain, A., Ross, A., Pankanti, S.:
\newblock Biometrics: A tool for information security.
\newblock IEEE Trans. on Information Forensics and Security \textbf{1} (2006)
  125--143

\bibitem{[Jain99Kluwer]}
Jain, A., Bolle, R., Pankanti, S., eds.:
\newblock Biometrics - Personal Identification in Networked Society.
\newblock Kluwer Academic Publishers (1999)

\bibitem{[Monro07]}
Monro, D., Rakshit, S., Zhang, D.:
\newblock {DCT-B}ased iris recognition.
\newblock IEEE Trans. on Pattern Analysis and Machine Intelligence
  \textbf{29}(4) (April 2007)  586--595

\bibitem{[schneier99usesAbuses]}
Schneier, B.:
\newblock The uses and abuses of biometrics.
\newblock Communications of the ACM \textbf{48} (1999)  136

\bibitem{[Ratha01]}
Ratha, N., Connell, J., Bolle, R.:
\newblock An analysis of minutiae matching strength.
\newblock Proc. International Conference on Audio- and Video-Based Biometric
  Person Authentication, AVBPA \textbf{Springer LNCS-2091} (2001)  223--228

\bibitem{[Soutar99]}
Soutar, C., Gilroy, R., Stoianov, A.:
\newblock Biometric system performance and security.
\newblock Proc IEEE Workshop on Automatic Identification Advanced Technologies, AIAT (1999)

\bibitem{[Fierrez07]}
Fierrez, J., {Ortega-Garcia}, J., {Torre-Toledano}, D.,
{Gonzalez-Rodriguez},
  J.:
\newblock {B}io{S}ec baseline corpus: A multimodal biometric database.
\newblock Pattern Recognition \textbf{40}(4) (April 2007)  1389--1392

\bibitem{[Gonzalez]}
Gonzalez, R., Woods, R.:
\newblock Digital Image Processing.
\newblock Addison-Wesley (2002)

\bibitem{systemLiborB}
Masek, L., Kovesi, P.:
\newblock Matlab source code for a biometric identification system based on
  iris patterns.
\newblock The School of Computer Science and Software Engineering, The
  University of Western Australia (2003)

\bibitem{[Daugman04]}
Daugman, J.:
\newblock How iris recognition works.
\newblock IEEE Transactions on Circuits and Systems for Video Technology
  \textbf{14} (2004)  21--30

\bibitem{[DaugmanOnlineIrisLiveness]}
Daugman, J.:
\newblock Anti spoofing liveness detection.
\newblock {available on line at
  http://www.cl.cam.ac.uk/users/jgd1000/countermeasures.pdf}

\bibitem{[Pacut06]}
Pacut, A., Czajka, A.:
\newblock Aliveness detection for iris biometrics.
\newblock Proc. IEEE Intl. Carnahan Conf. on Security Technology, ICCST (2006)
  122--129

\end{thebibliography}

\end{document}